\documentclass[a4paper]{article}
\usepackage{amsmath,graphicx}
\usepackage{diagbox}
\usepackage{multirow}
\usepackage{cite}
\usepackage{indentfirst}
\usepackage[a4paper,left=3cm,right=3cm,top=2.5cm,bottom=2.5cm]{geometry}

\providecommand{\keywords}[1]
{
  \small	
  \textbf{\textit{Keywords---}} #1
}

\title{Neural Networks with Divisive normalization \\ for image segmentation with application in cityscapes dataset}

\author{Pablo Hernández-Cámara, Valero Laparra and Jesús Malo \\\\
Image Processing Lab., Universitat de València\\46980 Paterna, Spain\\
pablo.hernandez-camara@uv.es, valero.laparra@uv.es, jesus.malo@uv.es}

\begin{document}
%
\maketitle
\begin{abstract}
One of the key problems in computer vision is adaptation: models are too rigid to follow the variability of the inputs. 
The canonical computation that explains adaptation in sensory neuroscience is \emph{divisive normalization}, and it has appealing effects on image manifolds.
In this work we show that including \emph{divisive normalization} in current deep networks makes them more invariant to non-informative changes in the images. 
In particular, we focus on U-Net architectures for image segmentation. 
Experiments show that the inclusion of \emph{divisive normalization} in the U-Net architecture leads to better segmentation results with respect to conventional U-Net. The gain increases steadily when dealing with images acquired in bad weather conditions.
In addition to the results on the Cityscapes and Foggy Cityscapes datasets, we explain these advantages through visualization of the responses: the equalization induced by the \emph{divisive normalization} leads to more invariant features to local changes in contrast and illumination\footnote{An extended version is under consideration at Patter Recognition Letters.}.

\end{abstract}

\keywords{Divisive normalization, Adaptation, Manifold alignment, Image segmentation, Neural networks.}

\section{Introduction}
\label{sec:intro}

\begin{figure}[ht]
\begin{minipage}[b]{1.0\linewidth}
  \centering
  \centerline{\includegraphics[width=9cm]{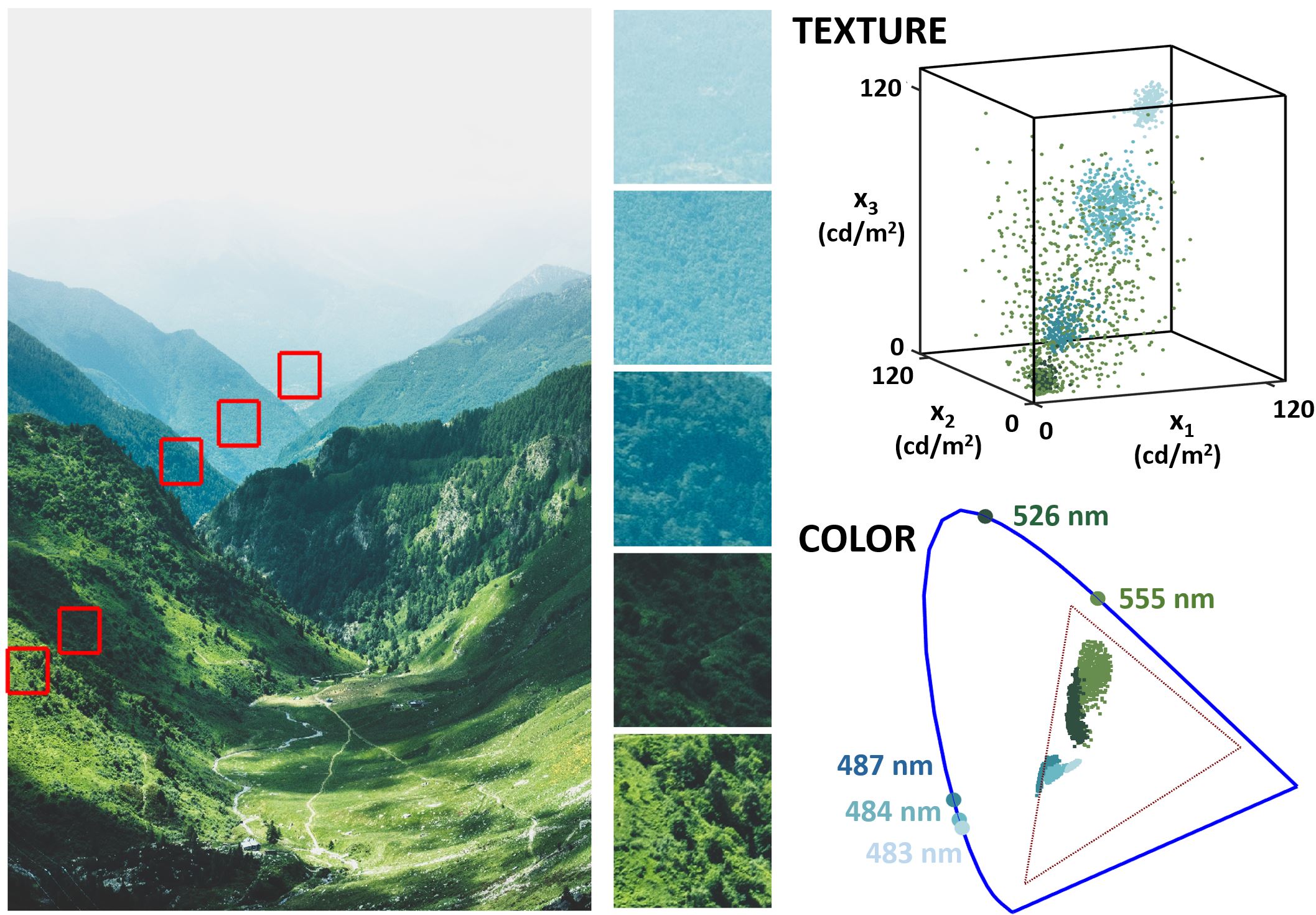}}
\end{minipage}
\caption{\textbf{The problem:} 
changes along the image due to non-informative 
factors. In this example, texture and color locally change due to the shadows, illumination, atmospheric scattering and depth (see highlighted regions).
Associated problems for statistical learning are shown in the scatter plots (where samples are colored according to their region).
The top-right plot shows changes in the spatial texture defined by the luminance of neighbor pixels: the contrast (distance from the diagonal) is smaller for lighter regions (away from the origin), and the energy in the different directions of the image space (spatial frequency) locally changes along the image. Atmospheric scattering implies local changes in the hue: see the shift in dominant wavelengths in the CIExy color diagram. }
\label{TheProblem}
\end{figure}

A fundamental problem in image analysis is the variability of the image manifold depending on acquisition conditions~\cite{Bengio09}. 
Usually sources are not stationary: the texture and color of an object can be different depending on the acquisition conditions (see Fig.~\ref{TheProblem}). This has an obvious negative impact on applications such as segmentation.
Adaptation to (and compensation of) non-informative factors of variation is key for optimal segmentation. 
However it is not clear how deep-learning, e.g. the popular U-Nets~\cite{ronneberger2015unet}, actually achieve this.

In this work we propose the use of
the canonical computation that accounts for adaptation in biological neurons, namely \emph{divisive normalization}~\cite{Carandini2012NormalizationAA}, to improve adaptation of artificial networks in an explainable way. We illustrate this adaptation ability in image segmentation under different atmospheric conditions.

\section{Why Using Divisive normalization?}
\label{sec:div_norm}

The linear+nonlinear structure of conventional artificial neurons~\cite{haykin2009neural} comes from the seminal  computation proposed for sensory neurons~\cite{HubelWiesel1959}:  
$\mathbf{x} \overset{\mathcal{L}}{\longrightarrow}  \mathbf{z} \overset{\mathcal{N}}{\longrightarrow} \mathbf{y}$, 
where the intermediate linear response, $\mathbf{z} = \mathcal{L} \cdot \mathbf{x}$, is given by the matrix $\mathcal{L}$, that contains the so called (linear) receptive fields, and original versions of $\mathcal{N}(\cdot)$ were simple point-wise threshold or saturating functions~\cite{Naka1966SpotentialsFL} eventually rectified. These were the inspiration for current sigmoids/ReLU in deep-learning~\cite{haykin2009neural}. 

In sensory neuroscience, the \emph{divisive normalization} model of $\mathcal{N}(\cdot)$ was a way to account for the inhibitory effect of neighbor neurons within a layer~\cite{Carandini2012NormalizationAA}:
\begin{eqnarray}
y_{ip} &=& \frac{z_{ip}}{\left( \beta_{i} + \sum_{jp'}{\gamma_{ijpp'}|z_{jp'}|^{\alpha_{j}}}\right)^{\epsilon_{i}}}
\label{GDN}
\end{eqnarray}
where the linear response of a neuron tuned to the $i$-th feature and to the $p$-th spatial location, $z_{ip}$, is inhibited (normalized) by a pool of the activity of neighbor neurons tuned to other features and spatial locations $z_{jp'}$. The constant $\beta_i$ determines the level in which the pool generates effective inhibition, and the exponents control the norm of the pool.
While the interaction kernel in the denominator, $\gamma$, can have whatever dense structure~\cite{martinez2018derivatives}, 
in visual neuroscience the spatial interaction is usually assumed to be convolutional in the fovea~\cite{Watson97}. 

\vspace{0.1cm}

\begin{figure}[t]
\begin{minipage}[b]{1.0\linewidth}
  \centering
  \centerline{\includegraphics[width=9cm]{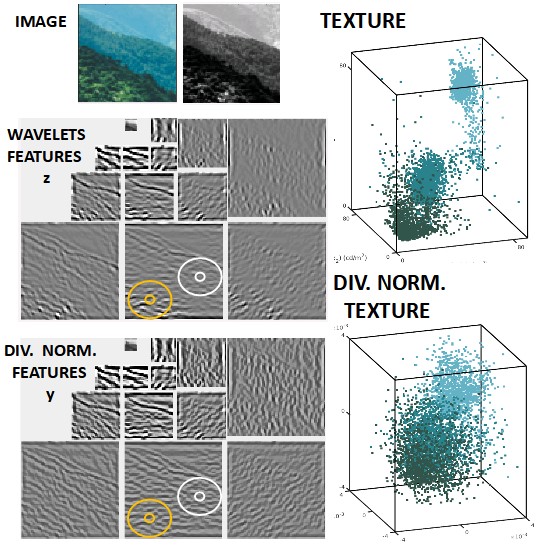}}
\end{minipage}
\caption{\textbf{The idea for the solution:} manifold equalization though  Divisive Normalization. 
At the top-left we have the luminance input to a linear+nonlinear model of the V1 cortex based on wavelets and divisive normalization~\cite{schwartz2001natural,martinez2018derivatives,Martinez19}. 
Below we have the wavelet-like linear features, $\mathbf{z} = \mathcal{L}\cdot\mathbf{x}$, and the corresponding divisive normalized features, $\mathbf{y} = \mathcal{N}(\mathbf{z})$.  The scatter plots display the distributions of image patches taken from $\mathbf{x}$ and $\mathbf{x'}$.
}
\label{TheSolution}
\end{figure}

However, why using this biological transform to improve artificial networks devoted to image segmentation?
There are examples of the effectiveness of \emph{divisive normalization} in many applications as image coding~\cite{Epifanio03,Malo2006NonlinearIR,balle2016end}, restoration~\cite{gutierrez2005regularization,laparra2017perceptually}, distortion metrics~\cite{pons1999image,Laparra_10,hepburn_2020}, classification~\cite{coen2013impact,Giraldo2019IntegratingFN,giraldo_schwartz_iclr2021,miller2022divisive} or even referring to its appealing statistical properties~\cite{schwartz2001natural,Malo06b,Malo10,balle2015density,Malo20}. Here we illustrate the effect of \emph{divisive normalization} with an explicit example.  Fig.~\ref{TheSolution} considers a patch from the forest (from Fig.~\ref{TheProblem}) that displays a space varying contrast due to atmospheric effects, and shows how \emph{divisive normalization} 
modifies the features and the image manifold in a positive way to identify this object class.  

The spatial texture problem is enough to stress the point of local normalization so in this motivating example we restrict ourselves to the luminance channel\footnote{Note that local color compensation could also be done using modern divisive normalization formulations of classical Von Kries ideas~\cite{Hillis_05,Fairchild13}.}. In this achromatic context, a reasonable physiological model of texture perception includes a linear layer of wavelet-like filters (in this illustration $\mathcal{L}$ is a steerable wavelet transform as in~\cite{schwartz2001natural,martinez2018derivatives}), and a \emph{divisive normalization} (in this case with the psychophysically tuned parameters in~\cite{Martinez19}).  

The neighborhood in $\gamma$ that defines the pooling region (illustrated in Fig.~\ref{TheSolution} by the circles in the feature vectors or wavelet bands) has a qualitatively relevant effect: the division by the local pool moderates the response in regions where contrast is high, in orange, while boosts the response in regions of low contrast, in white. See the difference between the features before and after the local normalization, and note the equalization of the amplitude of the edges along the field of view. The local normalization compensates for the contrast variation along the image due to atmospheric conditions.
As a result, while the distribution of the samples from the original luminance channel display separate clusters corresponding to spatial variations of the texture, the different samples have been compacted due to the contrast equalization effect. 

In the discussion we will explicitly check if the artificial networks for segmentation equiped with divisive normalization also develop this contrast enhancement and equalization effects that have been found in biological neurons. 

\section{Models and Experiments}
\label{sec:models}


In order to confirm the intuition obtained from the 1-layer, purely biological, non-optimized model presented in the previous section, 
here we implemented the generic \emph{divisive normalization} layer in Eq.~\ref{GDN} in an automatic differentiation context. 
Therefore, one may include our layer at any place of any architecture and optimize the network for the desired task. 

Qualitatively critical points are (a) the nature of the interaction kernel, $\gamma$, and (b) the specific locations to include the normalization in the U-Nets for segmentation.

Regarding the nature of $\gamma$, the above example points out that the consideration of spatial neighborhoods is convenient to get the local contrast equalization along the visual field required to overcome shadow, scattering or fog. 
The recent literature that exploits automatic differentiation  
shows a range of kernel structures: some do not consider spatial interactions 
(either in a dense~\cite{balle2015density,balle2016end,hepburn_2020} or convolutional~\cite{miller2022divisive} combinations of features); while others do with some restrictions (either uniform weights~\cite{ren2017normalizing}, a ring of locations~\cite{Giraldo2019IntegratingFN,giraldo_schwartz_iclr2021}, or special symmetries in the space~\cite{burg2021learning}).
Following biology~\cite{Watson97,schwartz2001natural,Carandini2012NormalizationAA,Martinez19} and the intuition pointed out in the previous section. In our experiments we use the version in Eq.~\ref{GDN} which includes a general spatial surround, as in~\cite{ren2017normalizing}, but allowing variable weights over space. Additionally a general (dense) interaction over channels was considered. Besides the $\gamma$ parameter, the $\beta$ parameter is also trained, but we set $\alpha$ and $\epsilon$ to $1$ which we found it makes the training more stable.

%

\begin{figure}[t]
\begin{minipage}[b]{1.0\linewidth}
  \centering
  \centerline{\includegraphics[width=9cm]{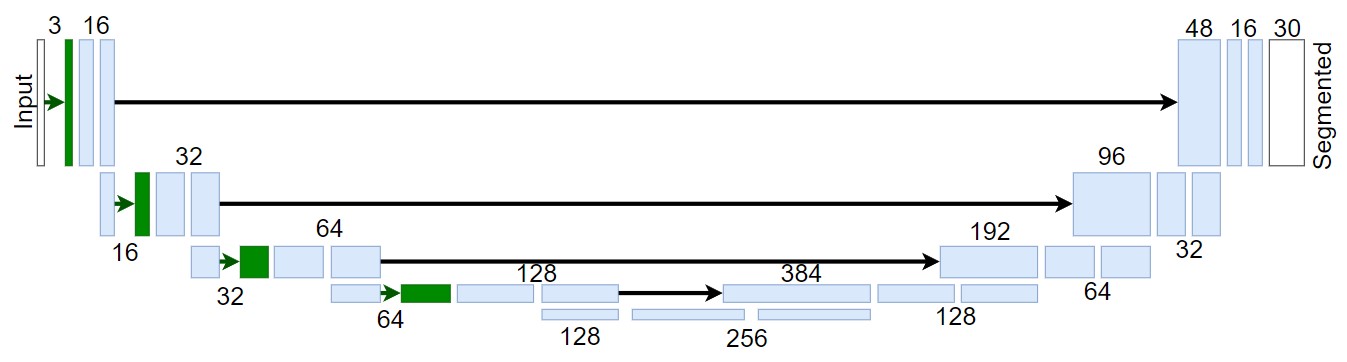}}
\end{minipage}
\caption{U-Net for segmentation including 4 DN layers (in green). Black arrows represent the skip unions. The model with only one DN layer had only the first green layer. The model without DN layers did not have any of the green layers. The consideration of the DN layers only increases the number of parameters by an 1.8$\%$.}
\label{fig:segmentation_model_4_gdn}
\end{figure}

Regarding where to include the proposed normalization in U-Nets, we recall the 
\emph{color} and \emph{texture} problems pointed out in Fig.~\ref{TheProblem}:
following color appearance models~\cite{Hillis_05,Fairchild13}, constancy maybe 
addressed by a single normalization of the photoreceptors (right before the first convolution), while texture equalization over space may require normalization at deeper stages where 
filters tuned to specific patterns have emerged.
This intuition lead us to the specific scheme in Fig.~\ref{fig:segmentation_model_4_gdn} where \emph{green} layers stand for the proposed \emph{divisive normalization}.
Note that we propose modifications only in the \emph{encoding/compressive} part, where the features that will lead to the segmentation are computed. The color/texture distinction shapes the experiments because we will compare 
the behavior of the improved network (4 DN layers) with regard to the No-DN network, but we will also consider a 1 DN layer network eventually addressing only the color problem, but not the texture problem.

In the experiments we used the Cityscapes dataset \cite{Cordts2016Cityscapes}, that includes scenes with annotated segmentation ground truth and the Foggy Cityscapes dataset \cite{fog_cityscapes}, that includes degraded versions of the scenes simulating poor weather conditions of controlled (low, medium, high) severity. In fog, spatial varying degradation is a major problem. The bottom pixels of each image were cropped from $1024 \times 2048$ to $768 \times 2048$ and then normalized and resized to $96 \times 256$ to maintain the aspect ratio. We used 2675, 300 and 500 images for train, validate and test. 
We used \textit{MAE} loss function, a batch-size of 64 and Adam optimiser with an initial learning rate of $0.001$ during 500 epochs with the original scenes (in regular weather conditions). We kept the models with higher Intersection over Union (IoU) in validation (in regular weather) and used these for the test (in poor weather). 
Testing in more general datasets where visibility is strongly reduced will confirm or refute the intuition in Fig.~\ref{TheSolution}.

\section{Results and Discussion}
\label{sec:results}

%

\begin{table}[ht]
\begin{center}

\begin{tabular}{|c||c|c|c|}
\hline
Dataset & No DN & 1 DN & 4 DN \\ \hline\hline
Original & 0.73 & 0.73 (0.0\%) & 0.78 (6.8\%) \\ \hline
Low fog & 0.65 & 0.63 (-3.1\%) & 0.70 (7.7\%)  \\ \hline
Middle fog & 0.55 & 0.52 (-5.5\%) & 0.61 (10.9\%)  \\ \hline
High fog & 0.39 & 0.42 (7.7\%) & 0.46 (17.9\%)  \\[0.15cm] \hline
\end{tabular}

\begin{tabular}{|c||c|c|c|}
\hline
\multicolumn{4}{|c|}{IoU reductions (due to fog)} \\ \hline
Dataset change & No DN & 1 DN & 4 DN \\ \hline\hline
Original-low fog & -11.0\% & -14.3\% & -9.8\% \\ \hline
Original-middle fog & -24.9\% & -28.8\% & -21.9\% \\ \hline
Original-high fog & -46.9\% & -42.6\% & -40.9\% \\ [0.15cm] \hline
\end{tabular}
\caption{\textbf{Top:} IoU performance in test. Improvements of the use of DN layers with regard no using DN in parenthesis.
\textbf{Bottom:} Reductions in test IoU when comparing each fog level with the original.}
\label{AllResults}
\end{center}
\end{table}

The top panel in Table \ref{AllResults} shows the test results of each model in the four weather conditions for models trained only using the original (good weather conditions) images.  
The bottom panel shows how the test IoU changes in poor conditions with regard to regular weather. 

The first observation is that using 4-DN layers gives substantial  increase in IoU in all cases. Using just 1-DN only helps in really heavy fog. This is consistent with the previous explanation since introducing fog into the images mainly changes the textures, which as previously mentioned requires several normalizations in deep stages.
Second, as expected, for progressively heavier fog, IoU gets reduced in all cases.  
However the conventional architecture is more sensitive to the decrease in visibility (bigger reductions in performance) than the architecture with 4-DN layers.
And third, the use of 4-DN always leads to improvements with regard the no-DN case, but it is important to see that the gains get progressively bigger when the acquisition conditions are poor.



\begin{figure*}[t!]
\begin{minipage}[b]{1.0\linewidth}
  \centering
  \centerline{\includegraphics[width = 16.8cm, height = 7.8cm]{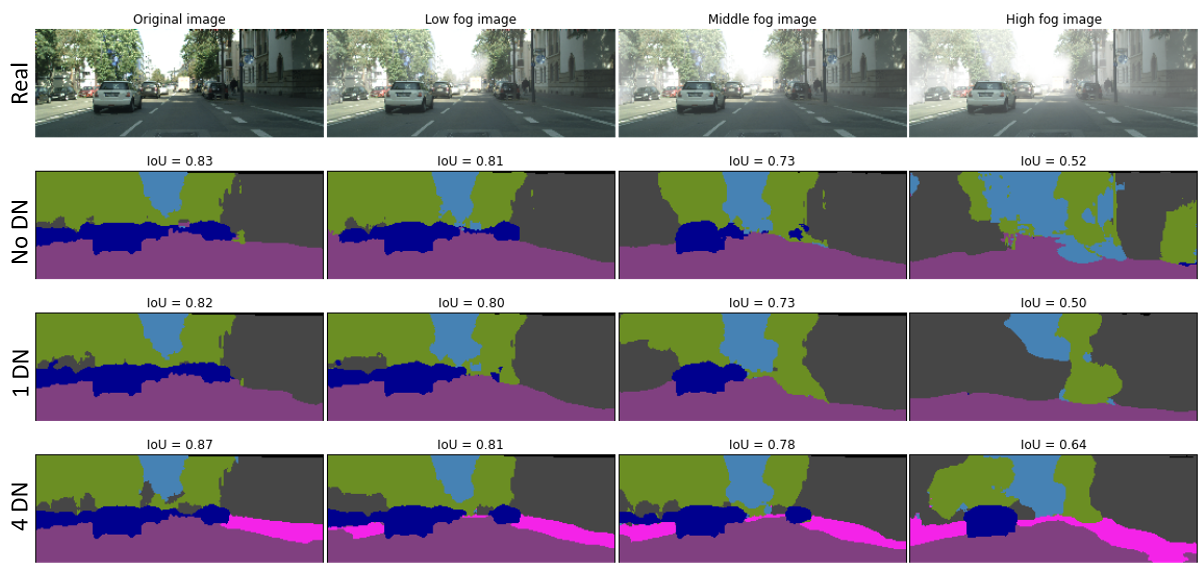}}
\end{minipage}
\caption{Segmentation results for the four datasets (original, low, middle and high fog) and the  three models.}
\label{fig:Segmentation_results}
\end{figure*}

\begin{figure*}[ht!]
\begin{minipage}[b]{1.0\linewidth}
  \centering
  \centerline{\includegraphics[width = 16cm, height = 7cm]{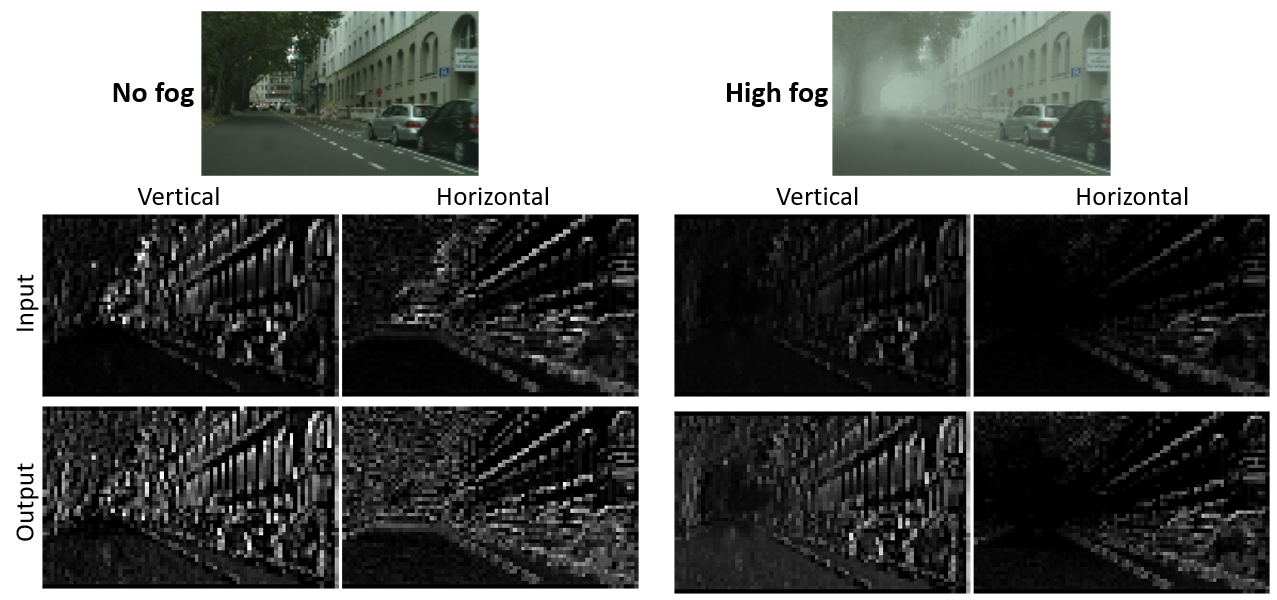}}
\end{minipage}
\caption{Effect of the 2nd-DN layer in two channels, one tuned to vertical and one tuned to horizontal features: activity before (input) and after (output) the divisive normalization layer. Results are shown for two different fog situations.}
\label{fig:channels_Second_dn}
\end{figure*}



Figure \ref{fig:Segmentation_results} shows illustrative examples of the predictions, which are consistent with average IoUs in the tables.  
Note that only the 4-DN model is able to get the border in the shadow and preserve the detection of the car in heavy fog.



In order to understand the above advantages we explored the effect of the two initial DN layers. The first DN, applied to the input RGB signal performs a Weber-like saturating non-linearity (not shown), very much like a square root equalization in the separate channels: it boosts low values and moderates high values. 
More interestingly, Fig.~\ref{fig:channels_Second_dn} shows the effect of the second DN layer on two illustrative features tuned to horizontal and vertical edges (key for object recognition). We see that in the foggy image the deepest parts of the image were impossible to distinguish. However, after applying the DN transformation that the model had learnt with the original (no foggy) images, the output is much more like the case without fog, now being able to distinguish details that are further away.


\vspace{-0.25cm}

\section{Conclusion}
\label{sec:conclusion}

We presented results in image segmentation when including \emph{divisive normalization} in the U-Net architecture. 
We found that introducing more than one \emph{divisive normalization} layer leads us to an improvement in IoU in the Cistyscapes dataset. Moreover, a model trained only with images acquired in good weather conditions obtains an increase in IoU over conventional U-Net when tested with images in bad weather conditions. In fact, the higher the level of fog introduced in the images, the higher the advantage, which shows that this transformation helps the models to be more invariant under the non-informative variations introduced by contrast and luminance changes.


\vspace{-0.25cm}

\bibliographystyle{IEEEbib}
\bibliography{strings,refs}

\end{document}